\documentclass[runningheads]{llncs}
\usepackage{graphicx}
\usepackage{amsmath}
\usepackage{algorithmic}
\usepackage{graphicx}
\usepackage{textcomp}
\usepackage{xcolor}
\usepackage{enumitem}
\usepackage{hyperref}
\usepackage{mathabx}
\usepackage{subcaption}
\usepackage{multirow}
\usepackage{mdframed}
\usepackage{cite}

\mdfdefinestyle{RQFrame}{
 outerlinewidth=0pt,
 skipabove=0pt,
 skipbelow=0pt,
 innertopmargin=6pt,
 innerbottommargin=0pt,
 linewidth=0pt,
 topline=false,
 rightline=false,
 leftline=false,
 innerrightmargin=4pt,
 innerleftmargin=4pt}

\newcounter{RQCounter}

\definecolor{Gray}{gray}{0.9}

\newboolean{showcomments}
\setboolean{showcomments}{true} % toggle to show or hide comments
\ifthenelse{\boolean{showcomments}}
  {\newcommand{\nb}[2]{
    \fcolorbox{Gray}{yellow}{\bfseries\sffamily\scriptsize#1}
    {\sf\small$\blacktriangleright$\textit{#2}$\blacktriangleleft$}
   }
   
  }
  {\newcommand{\nb}[2]{}
   
  }

\begin{document}
\title{The Analysis of Online Event Streams: Predicting the Next Activity for Anomaly Detection}
\titlerunning{Predictive Anomaly Detection for Online Event Streams}
% If the paper title is too long for the running head, you can set
% an abbreviated paper title here
%
\author{Suhwan Lee\inst{1} \and
Xixi Lu\inst{1} \and
Hajo A. Reijers\inst{1}}
\authorrunning{S. Lee et al.}
% First names are abbreviated in the running head.
% If there are more than two authors, 'et al.' is used.
%
\institute{Utrecht University,
Utrecht, The Netherlands \\
\email{\{s.lee, x.lu, h.a.reijers\}@uu.nl}
}
\maketitle              % typeset the header of the contribution
\begin{abstract}
% Successful business process analysis requires high event log quality to be correctly aware of recorded event data and improve process execution. Traditional anomalies detection in process mining has focused on offline event log. While a few approaches have been applied on online streaming events, majorities aim at filtering deviated cases from the normal process behavior. The literature propose statistical analysis methods which is based on probability of executed processes. Machine learning approaches may outperform statistical analysis because of the complex interrelation between events and dynamic process change on streaming events. In this paper, we propose a framework to detect anomalous events from online process execution using machine learning and deep neural network. We present evaluation with various well-known models and effective anomalous event detection methods for online event log. 

% My first try. 
Anomaly detection in process mining focuses on identifying anomalous cases or events in process executions. The resulting diagnostics are used to provide measures to prevent fraudulent behavior, as well as to derive recommendations for improving process compliance and security.
Most existing techniques focus on detecting anomalous cases in an offline setting. However, to identify potential anomalies in a timely manner and take immediate countermeasures, it is necessary to detect event-level anomalies online, in real-time.   
In this paper, we propose to tackle the online event anomaly detection problem using next-activity prediction methods. More specifically, we investigate the use of both ML models (such as RF and XGBoost) and deep models (such as LSTM) to \emph{predict} the probabilities of next-activities and consider the events predicted unlikely as anomalies. We compare these predictive anomaly detection methods to four classical unsupervised anomaly detection approaches (such as Isolation forest and LOF) in the online setting. Our evaluation shows that the proposed method using ML models tends to outperform the one using a deep model, while both methods outperform the classical unsupervised approaches in detecting anomalous events. 

% detecting anomalies at the event level in an online setting pose new challenges such as having limited training data or requiring to update the model frequently. 

%  their performance and suitability for detecting anomalous events in streaming data.

% Interestingly, our results show that the proposed predictive detection using classical ML models actually tends to outperform a LSTM model, while they both outperform the classical unsupervised approaches, in online anomalous event detection.

% Interestingly, our results show that the proposed predictive detection using classical ML models actually tends to outperform a LSTM model in online anomalous event detection as well as the classical unsupervised approaches .

\end{abstract}
\section{Introduction}
\label{sec:Introduction}
% TODO: what is current status
Information systems, empowered by blockchain~\cite{guo2018blockchain} and IoT systems~\cite{kolozali2014knowledge}, allow an enormous amount of event data to be generated and logged in real-time. The data analytic techniques, such as process mining, are developed to manage the big volume of recorded real-time data. Process mining is a technique to identify and acknowledge the recorded events and gain insights to improve process execution~\cite{van2016data}. Recently, process mining has focused on process management and analysis on online settings including process discovery~\cite{burattin2012heuristics}, conformance checking~\cite{burattin2017framework}, and process monitoring techniques~\cite{maisenbacher2017handling}.

% TODO: why online event outlier detection. 
Anomaly detection in process mining aims to detect anomalous behavior in event data~\cite{ghionna2008outlier,DBLP:conf/bpm/LuFBA15}.
Such techniques have been used to identify potential fraudulent behavior to prevent compliance violations~\cite{burattin2017framework}. In addition, they are also used to detect log quality issues to improve data quality~\cite{nguyen2019autoencoders}. Most existing unsupervised anomaly detection techniques focus on the \emph{case-level} or are situated in \emph{offline} settings~\cite{ghionna2008outlier}, i.e., they take as input of a batch event log that contains a set of completed cases.

In practice, detecting anomalies in \emph{online} streaming settings has many advantages, such as being able to take action and timely counter measures. Timely detection also helps dealing with concept drift. At the same time, online anomaly detection also faces many challenges. Unlike an offline setting, which only deals with completed cases, online detection should be able to continuously handle incomplete, ongoing cases. Moreover, online detection should pinpoint anomalies at \emph{event-level} to allow timely, concrete reactions. For example, if credit card fraud is established, the techniques should immediately detect and pinpoint which purchase events are suspicious.

In this paper, we propose to tackle unsupervised anomalous event detection by predicting which activity is next, assuming an ongoing case.
More specifically, we first learn a predictive model to predict next activities by preprocessing the completed cases into feature vectors and using machine learning-based classification algorithms (such as Random Forest, XGBoost, LSTM). When a new event of an ongoing case arrives, we apply that predictive model to predict the possible activities and their probabilities using the previous events. The less likely that an activity occurs, the more likely it is an anomaly. 

We conduct an evaluation and compare our approach to other approaches that simply encode the events and apply unsupervised anomaly detection algorithms (including Isolation Forest, LOF, OCSVM) and show that our approach performs better in terms of F1 scores. Therefore, this seems a promising direction for online anomaly detection, which raises many new research challenges.%\xixi{Can you check whether this is true} 

% The contributions :
% (1) Instead of only using LSTM,  We generalize the idea and propose an approach that may use any next-activity prediction method (including RF, XGBoost, ..).
% (2) We investigate the difference between classical ML models and deep models and their suitability for streaming event data, 
% (3) we compare our generalized predictive anomaly detection approach to unsupervised anomaly detection approaches. 

The paper is organized as follows. Related work is discussed in the next section. Section~\ref{sec:preliminaries} describes the preliminary knowledge for easy understanding on the proposed method. Section~\ref{sec:methodology} introduces the proposed method and setup of evaluation is described in Section~\ref{sec:empirical_evaluation}. The experimental results are reported in Section~\ref{sec:results}, while conclusions and challenges are drawn in Section~\ref{sec:conclusion}.

\section{Related work}
\label{sec:related_works}
% XL: explain your overview

In this section, we discuss event anomaly detection techniques, that are related to our approach. See Table~\ref{tab:related_works_table}. We categorize existing approaches along two dimensions, (1) \emph{offline} versus \emph{online} and (2) \emph{case-level} versus \emph{event-level}. We discuss them accordingly. Then, our contributions in relation to related works are presented.

% Please add the following required packages to your document preamble:
% \usepackage{multirow}
% \usepackage{graphicx}
\begin{table}[!t]
\centering
\caption{Comparison of related anomaly detection approaches}
\label{tab:related_works_table}
\resizebox{\textwidth}{!}{%
\begin{tabular}{lccc}
\hline
 & \textbf{Online/Offline} & \textbf{Target status} & \textbf{Algorithm} \\ \hline
\multicolumn{1}{l|}{Ghionna et al.~\cite{ghionna2008outlier}} & \multicolumn{1}{c|}{\multirow{6}{*}{Offline}} & \multicolumn{1}{c|}{\multirow{3}{*}{Case-level}} & Markov cluster \\ \cline{1-1} \cline{4-4} 
\multicolumn{1}{l|}{Sani et al.~\cite{sani2017improving}} & \multicolumn{1}{c|}{} & \multicolumn{1}{c|}{} & Occurance proability \\ \cline{1-1} \cline{4-4} 
\multicolumn{1}{l|}{Khatuya et al.~\cite{khatuya2018adele}} & \multicolumn{1}{c|}{} & \multicolumn{1}{c|}{} & Ridge regression \\ \cline{1-1} \cline{3-4} 
\multicolumn{1}{l|}{Nguyen et al.~\cite{nguyen2019autoencoders}} & \multicolumn{1}{c|}{} & \multicolumn{1}{c|}{\multirow{3}{*}{Event-level}} & Autoencoder \\ \cline{1-1} \cline{4-4} 
\multicolumn{1}{l|}{Savickas and Vasilecas~\cite{savickas2018belief}} & \multicolumn{1}{c|}{} & \multicolumn{1}{c|}{} & Bayesian belief network \\ \cline{1-1} \cline{4-4} 
\multicolumn{1}{l|}{Nolle et al.~\cite{nolle2022binet}} & \multicolumn{1}{c|}{} & \multicolumn{1}{c|}{} & LSTM \\ \hline
\multicolumn{1}{l|}{Tavares et al.~\cite{tavares2019overlapping}} & \multicolumn{1}{c|}{\multirow{4}{*}{Online}} & \multicolumn{1}{c|}{\multirow{3}{*}{Case-level}} & Process model conformity check \\ \cline{1-1} \cline{4-4} 
\multicolumn{1}{l|}{Ko and Comuzzi~\cite{ko2020online}} & \multicolumn{1}{c|}{} & \multicolumn{1}{c|}{} & Leverage score calculation \\ \cline{1-1} \cline{4-4} 
\multicolumn{1}{l|}{Neto et al.~\cite{vertuam2021use}} & \multicolumn{1}{c|}{} & \multicolumn{1}{c|}{} & Autocloud \\ \cline{1-1} \cline{3-4} 
\multicolumn{1}{l|}{Van Zelst et al.~\cite{van2018filtering}} & \multicolumn{1}{c|}{} & \multicolumn{1}{c|}{Event-level} & Automaton processor \\ \hline
\end{tabular}%
}
\end{table}
% %%%%%%%%%%%%%%%%%%%%%%%%%%%%%%%%%%%%%
% Case level offline Anomaly detection algorithm %
% %%%%%%%%%%%%%%%%%%%%%%%%%%%%%%%%%%%%%

The classical approach to detect anomalies in process data is aimed at analyzing an offline event log, which has a fixed number of events. 
Regarding \emph{offline}, \emph{case-level} anomaly detection algorithms, existing works try to discover anomalous cases in an event log, which have infrequent patterns~\cite{ghionna2008outlier}\cite{sani2017improving} or statistically deviate to an event log~\cite{khatuya2018adele}. Sani et al.~\cite{sani2017improving} suggested an outlier filtering method based on observed subsequence, in which the activity below the threshold is classified as outlier according to the succeeding activity probability. Khatuya et al.~\cite{khatuya2018adele} proposed to use ridge regression to estimate an anomaly score of the individual case by obtaining a statistical distribution of event log features. Although some events in the anomalous case may be normal, these methods are focused on the case rather than individual events.

% Regarding \emph{offline}, \emph{case-level} anomaly detection algorithm, Ghionna et al.~\cite{ghionna2008outlier} have proposed case level outlier detection techniques based on frequent execution patterns via Markov cluster algorithm. The infrequent process which is not listed on the obtained common pattern is classified as an outlier. Sani et al.~\cite{sani2017improving} have proposed an case level outlier filtering method based on observed subsequence, in which the activity below the threshold is classified as outlier according to the succeeding activity probability. Khatuya et al.~\cite{khatuya2018adele} have proposed to use ridge regression to estimate an anomaly score of the individual case by obtaining a statistical distribution of event log features. The introduced

% %%%%%%%%%%%%%%%%%%%%%%%%%%%%%%%%%%%%%
% Event level offline Anomaly detection algorithm %
% %%%%%%%%%%%%%%%%%%%%%%%%%%%%%%%%%%%%%

In case of \emph{offline}, \emph{event-level} anomaly detection, a probabilistic process model has been proposed~\cite{savickas2018belief}. Savickas and Vasilecas used Bayesian Belief Networks to detect an anomalous event. An event with a low probability according to the obtained table is classified as an outlier. To improve detector performance, some methods use a deep neural network~\cite{nguyen2019autoencoders, nolle2022binet}. Nguyen et al.~\cite{nguyen2019autoencoders} have proposed anomalous event detection and reconstruction framework, which uses Autoencoder to extract normative features from an event log. Nolle et al.~\cite{nolle2022binet} recently introduced a framework to detect case and event anomalies by obtaining the probability of the next event with the deep neural network LSTM. The model based on deep neural networks outperforms existing anomaly detection techniques. However, the works that detect anomalous events assume a constant process distribution in an event log. 

% The classical approach to detect anomaly in process data is aimed at analyz-ing an offline event log, which has a fixed number of events. Regardingoffline,case-levelanomaly detection algorithm, existing works try to discover anomalouscases in an event log, which have an infrequent patterns [4][17] or deviated sta-tistical distribution to an event log [7]. Sani et al. [17] have proposed an outlierfiltering method based on observed subsequence, in which the activity below thethreshold is classified as outlier according to the succeeding activity probability.Khatuya et al. [7] have proposed to use ridge regression to estimate an anomalyscore of the individual case by obtaining a statistical distribution of event logfeatures. These methods are focused on the cases rather than individual events.However, some of events in the anomalous case may be normal.

% %%%%%%%%%%%%%%%%%%%%%%%%%%%%%%%%%%%%%
% Case level online Anomaly detection algorithm %
% %%%%%%%%%%%%%%%%%%%%%%%%%%%%%%%%%%%%%

Regarding \emph{online}, \emph{case-level} anomaly detection, statistical leverage and data clustering models are used. Ko and Comuzzi~\cite{ko2020online} have proposed to use a sliding window with a recent event feature vector to calculate the statistical leverage score of the coming trace. When the sliding window is updated, a trace with a higher leverage score is classified as anomalous. Neto et al.~\cite{vertuam2021use} have described how to use Autocloud to detect anomalous cases from a stream of events. The model updates the data cluster and classifies anomalous data which is deviated from the existing cluster. To provide a process model, Tavares et al.~\cite{tavares2019overlapping} have proposed a framework that classifies an anomalous case by checking conformity between discovered model and the target case. The proposed methods are able to detect outliers in a streaming event log with a normative process change. Nevertheless, a streamed event is not simultaneously classified and the works are limited in taking into account case-level detection.

% Regarding \emph{online}, \emph{case-level} anomaly detection, Tavares et al.~\cite{tavares2019overlapping} have proposed a streaming events analysis framework, which also covers process discovery and process enhancement. The framework classifies anomalous case by checking conformity between discovered model and the target case. They could generate process model based on the trace distance cluster and trace frequency. Ko and Comuzzi~\cite{ko2020online} have proposed to use sliding window with recent event feature vector to calculate statistical leverage score of coming trace. When the sliding window is updated, a trace with higher leverage score is classified as anomalous. Neto et al.~\cite{vertuam2021use} have proposed to use Autocloud to detect anomalous case from a stream of event, specifically clustering data in unsupervised learning with a single hyperparameter.

% %%%%%%%%%%%%%%%%%%%%%%%%%%%%%%%%%%%%%
% Event level online Anomaly detection algorithm %
% %%%%%%%%%%%%%%%%%%%%%%%%%%%%%%%%%%%%%

Specifically for \emph{online}, \emph{event-level} anomaly detection, Van Zelst et al.~\cite{van2018filtering} have proposed automaton based filtering that uses a sliding bucket with a finite number of events. The model learns the probability of activity sequences. The advantage of this approach is that the model classifies a streaming event as soon as the event arrives. However, the detector performance may be limited, considering that the training data takes only activity occurrence and only one or two consecutive event sequence is used.

In summary, we see an opportunity in online, event-level anomaly detection to deal with process change and possible performance improvements. The traditional anomaly detection methods are not designed to cope with a streaming event log. The detector assumes a steady process and is not updated. In the case of anomaly detection in a streaming event log, existing works have mainly focused on case-level detection that are capable of catching anomalous data after the case is finished. In addition, the work for event-level detection uses a simple probabilistic model. In this paper, we propose an approach for online anomalous event detection. The proposed approach is based on the machine learning model for performance. The arrived event is classified by a retrained model taking into account possible process change.

% \section{Challenges}
% \label{sec:research_challenge}
% \input{files/research_challenge}

\section{Preliminaries}
\label{sec:preliminaries}
Before discussing the steps of the proposed approach in detail, let us explain some required preliminaries. This involves notations for the event log, the key components of pre-processing, and some detection mechanisms. The concepts discussed in this section are implemented to develop our anomaly detection approach, which is discussed in Section~\ref{sec:methodology}.

% %%%%%%%%%%%%%%%%%%%%%%%%%%%%%%%%%%%%%
% Explanation on notation of event log%
% %%%%%%%%%%%%%%%%%%%%%%%%%%%%%%%%%%%%%
\subsection{Event log}

An event log contains cases, which consist of a sequence of events. An event consists of multiple attributes including case id, activity, and timestamp. Let $\mathcal{G}_{att} = \{\mathcal{D}_1, \mathcal{D}_2, ..., \mathcal{D}_n \}$ be a set of all possible attributes and $\mathcal{D}_i$ be a set of all possible values for the attribute $i$. Attributes could be numerical or continuous values. For example, a timestamp takes a numerical value within an interval from beginning to end of an event log. The categorical attribute takes a value from a given set, e.g., an activity with a string data type is assigned within the labels $\{a_1, a_2, ..., a_n\}$. Therefore, we can express an event as a tuple $e=\langle c, act, tst \rangle$, where $c$, $act$, $tst$ are case id, one label from set of activity, and a point of time in an event log for timestamp, respectively.

\subsection{Next activity prediction}
\label{sec:prelim.next_activity_prediction}
% Function for Offline next activity prediction not anomaly detection
% Input and output
% The anomaly detection approach using next activity prediction classifies arrived event of running case as normal or anomalous with output matrix obtained from trained model and target event. The output matrix of next activity prediction consists of possible activity labels and probability of the candidates. The target event is classified as normal if the actual activity is listed on the possible activity dictionary and satisfies sufficient probability level, above \textit{anomaly threshold}, figured in the output matrix. As an example, \textit{e6} event in Fig.~\ref{fig:framework_proposed} arrives as the second event of case \textit{3} and has an activity label as \textit{D}. According to the output matrix, this event is classified as anomalous due to the low probability compared to a threshold. 

Next activity prediction is one of the techniques in process mining to predict a following activity of running case. Predictions are made using a \textit{classifier} that takes a fixed number of independent features as input. The classifier learns mathematical function to estimate target variables. This means that a classifier extracts features and predicts a probability of following activity from previous events of a case. The data in an event log and previous events of a target case are used as input for training and testing of the classifier, respectively. Both input data are pre-processed and encoded to a feature vector of equal size. The output from the classifier consists of possible activity labels and probability of the candidates. The activity with the highest probability is selected as a prediction.

\subsection{Unsupervised anomaly detection}
\label{sec:prelim.unsupervised}

% Function for how to classify normal or anomalous events
% Using inputs
% Input and output

The models used in unsupervised anomaly detection extract information from the data and map input matrix to a feature space. During a classifier training phase, the data in an event log is transformed and allocated to feature space. The classifier takes events from a running case, including target event, and calculates a distance between the running case and training data mapped into the feature space. If the target event sufficiently deviates from the training data, above the \textit{anomaly threshold}, the event is denoted as anomalous. Otherwise, the event is normal.

\section{Approach}
\label{sec:methodology}
% %%%%%%%%%%%%%%%%%%%%%%%%%%%%%%%%%%%%%
% Detail step of pedictive anomaly detection %
% %%%%%%%%%%%%%%%%%%%%%%%%%%%%%%%%%%%%%

This section presents in detail the proposed approach for detecting anomalous activities in streaming event logs: \emph{predictive anomaly detection}. The steps of the proposed approach are shown in Fig.~\ref{fig:framework_proposed}. We first explain our approach in an off-line setting in Section~\ref{subsec:pad}. Next, we discuss how the approach can be adapted in an online setting in Section~\ref{subsec:sliding_window}. %\xixi{Please check these references}

\subsection{Predictive Anomaly Detection (PAD)}
\label{subsec:pad}

\begin{figure}[!h]
        \centering
        \includegraphics[width=0.9\linewidth]{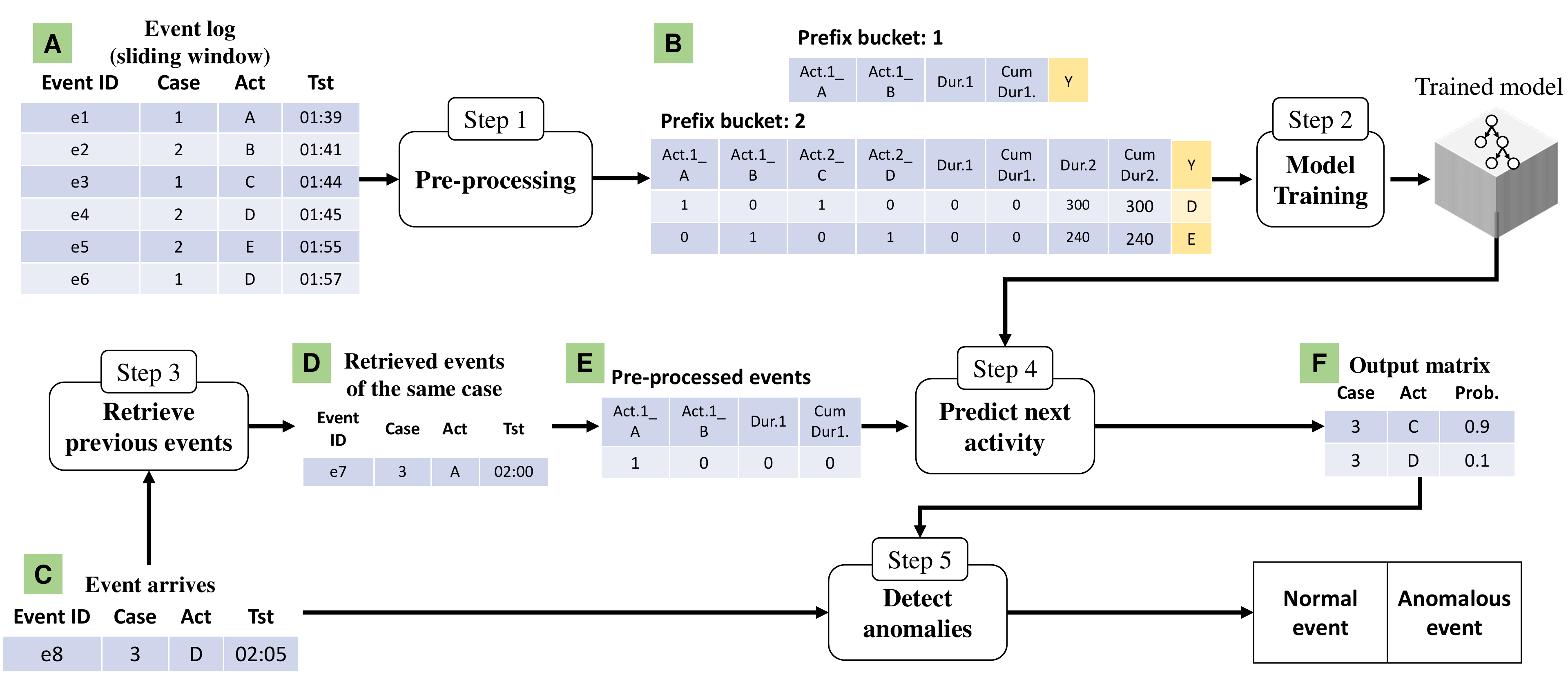}
        \caption{Online anomalous event detection procedure: Proposed approach}
        \label{fig:framework_proposed}
\end{figure}

The proposed approach uses machine learning classification methods to predict next activities for detecting anomalous events. We divide the approach into five steps, as shown in Fig.~\ref{fig:framework_proposed}. In the first two steps, we \emph{pre-process} the event log and \emph{train} a model to predict next activity for an event. In the step 3-5, we classify an event $e$ to be an anomaly or not by first \emph{retrieving} its previous event $e'$. Next, using the trained model, we \emph{predict} the next activities of $e'$ and their probabilities. We then \emph{detect} whether the probability of $e$ is above a threshold. In the following, we explain the approach in depth.

% An event log is used as an input for our approach. In the pre-processing step, the cases are transformed into a feature vector. The trained model is used to predict the next activity of a running case. The output of this phase is a matrix with a probability distribution of candidate activities for the next event. In detection step, the output matrix is used as a rubric to classify the arrived event of running case as normal or anomalous depending on the obtained probability. 

\subsubsection{Step 1 and 2: Pre-processing and Model Training}
To deploy machine learning techniques, which requires feature space in regular format, we transform the finished cases into a training dataset. The dataset is suitable for model training by using \emph{prefix-bucketing} and \emph{feature encoding}. Since our approach relies on the output of a next activity prediction approach, pre-processing methods are adopted from existing predictive process monitoring approaches~\cite{teinemaa2019outcome}.
% \textcolor{red}{XXX TODO XXX mention that you are following existing next activity prediction approaches, and mention specifically if there are any differences.}

% \subsubsection{Prefix-bucketing}
% \label{subsec:preliminary.prefix_bucketing}

Cases used for model training may have a different sequence of events. A common choice is to group cases into buckets by the same prefix length and separately pre-process the bucket for individual anomalies detector~\cite{leontjeva2016complex}. Multiple classifiers are trained to learn features from respective bucket which contains cases. Prefix bucket $n$ contains events from first to $n$th event of each case.

For example, let us consider two cases in an event log, both of which have 3 events (see Fig.~\ref{fig:framework_proposed}A). Two prefix length buckets are obtained, one with all prefixes of length 1, and the other with all prefixes of length 2. For each event in the buckets, its next activity is used as output label, \textit{Y}, for training a classifier (Fig.~\ref{fig:framework_proposed}B). 

% \textcolor{red}{TODO - Add example: For example, given the log shown in Fig. 1, we obtain two prefix buckets, one with all prefixes of length 1, and another with all prefixes of length 2.}

% \subsubsection{Feature encoding}
% \label{subsec:preliminary.feature_encoding}

In order to train a classifier, the collected events in the buckets are required to be modified as feature vector with fixed size. For feature encoding, we transform the attributes (e.g., activity and timestamp) into suitable features with index-base encoding method to maintain the order of events. The categorical attribute, such as activity label, is encoded using a one-hot encoding scheme by considering the order of events~\cite{leontjeva2016complex}. An encoded activity label $a$ for case $c$ at event $i$ is: 

\[  c_{i,act}=
		\begin{cases}
          1 &   \quad \text{if } c_{i,a} = Act.i\_a\\    
          0 &   \quad \text{Otherwise}
		\end{cases}
\]

Regarding a timestamp attribute, a point of the event occurrence is transformed into duration and cumulative duration of an event and a case, respectively (see Fig.~\ref{fig:framework_proposed}B). 
% [\xixi{Dur1 and CumDur1 are 0... Moreover, I just notice that the new case is a case 3. Why not event 2 of case 2?}] 
Event duration $Dur.i$ is elapsed time between a preceding event $i-1$ and a current event $i$, while cumulative duration $CumDur.i$ at event $i$ is aggregated event duration since the start of the case. Given an event timestamp, the obtained duration and cumulative duration respectively are:  $Dur.i = e_{tst, i} - e_{tst, i-1}$ and $CumDur.i = \sum_{n=1}^{i} Dur.n$ 

% \begin{align*}
%     Dur.i = e_{tst, i} - e_{tst, i-1}   \\
%     CumDur.i = \sum_{n=1}^{i} Dur.n
% \end{align*}

After prefix-bucketing and feature encoding, the pre-processed rows and columns are concatenated together as an input feature matrix for model training. 
% Unsupervised anomaly detection model takes this input matrix to train model, e.g., detecting model for $i$th event trained with input matrix from prefix bucket $i$. 

% \subsubsection{Next activity encoding}
% \label{subsec:preliminary.next_activity_encoding}

The objective of this phase is to learn a model (i.e., a classifier) that uses previous events to predict the activity of next event. 
In addition to the possible next activities, we also retrieve the probabilities of each possible activity from the model to determine how \emph{likely} or \emph{unlikely} the next activity is. 
% We wrap this prediction output of a trained model into a matrix that comprise the next activities and the probability of these activities.
% That is, the model learnt from input matrix built by prefix bucket $i$ is available to predict coming event activity $i$ of running case by taking prefix bucket $i-1$ as input data and $i$th event as target data for training phase, respectively. 
Note that we use the \emph{completed} cases for training a model. We then apply the model to predict the next coming event of \emph{running} cases in step 3-5.

\subsubsection{Step 3-5: Retrieve, Predict, and Detect}
We have trained a model that predicts the next possible activities. When a new event $e$ of a running case arrives (Fig.~\ref{fig:framework_proposed}C), we first retrieve its previous event $e'$ (Fig.~\ref{fig:framework_proposed}D) and the encoded event (Fig.~\ref{fig:framework_proposed}E). We use this encoded event as input for the trained model to predict the possible next activities and their probability, which we wrapped into an \emph{output matrix}(Fig.~\ref{fig:framework_proposed}F). 

We then use this output matrix to classify the new event $e$. 
If the activity of event $e$ is listed on the possible activities and its probability is above a sufficient level, i.e., the \textit{anomaly threshold}, this means that the target event occurs commonly after $e'$ and, therefore, is classified as normal. Otherwise, we classify $e$ as an anomaly. 

As an example, let us assume the anomaly threshold is 0.15. Event $e_8$ in Fig.~\ref{fig:framework_proposed} arrives as the second event of Case \textit{3} and has an activity label \textit{D}. We first retrieve the previous events of Case \textit{3}. We then apply the trained model to predict the possible next activities. According to the output matrix of the trained model, the probability of $e_8$ being activity $D$ is only 0.1. Due to this probability lower than the anomaly threshold, this event classified as anomalous. 

% %%%%%%%%%%%%%%%%%%%%%%%%%%%%%%%%%%%%%
% Sliding window & Retraining interval %
% %%%%%%%%%%%%%%%%%%%%%%%%%%%%%%%%%%%%%
\subsection{Online PAD using Sliding Window and Retraining Interval}
\label{subsec:sliding_window}
We have explained the proposed approach in an offline setting. In this section, we explain how the approach handles streaming events. 

A \emph{sliding window} is one approach to learn features from new observations of sequential data via updated window~\cite{hulten2001mining}. Before the pre-processing step, a sliding window collects a number of most recent completed cases from the streaming events and manages the data to be transformed. As a newly finished case from event streams arrives, a sliding window takes it as an input and places it at the beginning of the window. A case located at the end of the window is removed to maintain the fixed sliding window size ($W$). As long as $W$ is relatively small to the total number of cases, this procedure allows the machine learning model to be agilely retrained considering possible normal behavior changes of the dataset by collecting the information from more recent observations. However, a small sliding window size may be insufficient to properly train due to the lack of information on normal behavior. As well as issues on size for model training, a small sliding window size requires frequent retraining on the model, which leads to unnecessary recalculation of the outcome. We apply multiple parameters to investigate the sensitivity of window size to detect the optimal performance of the proposed approach in our evaluation.

Besides using the sliding window, we implement a parameter called \emph{retraining interval} $R$ to control the retraining frequency of the model. In essence, after retraining the model at a certain point in time $t$, we pause to retrain the model again until $R$ number of new cases are completed and updated in the sliding window. For example, if the absolute value of $R$ is $1$, then the model is retrained after each case is completed. If $R = W$, then the model is retrained after all the cases in the sliding window are updated. For such a retraining interval, we use a relative size to a sliding window, e.g., $R = 10\%$ of $W$. If the number of new cases inserted into the sliding window satisfies the retraining interval size condition, the anomaly detection model is retrained.

\begin{figure}[h]
        \centering
        \includegraphics[width=0.85\linewidth]{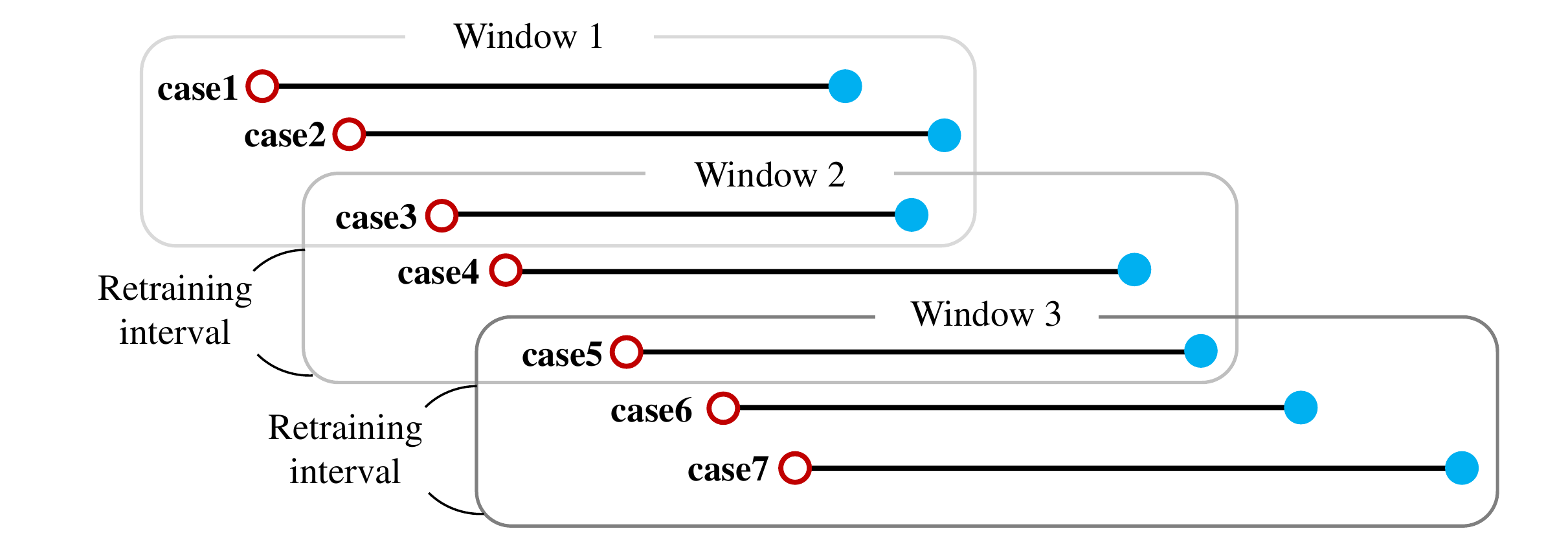}
        \caption{The architecture of sliding window and retraining interval}
        \label{fig:framework_sliding_window}
\end{figure}

Fig.~\ref{fig:framework_sliding_window} exemplifies how the sliding windows are updated along with a retraining interval. Let us assume there are 30 cases in an event log and 10\% of sliding window size with 66.7\% of retraining interval size. The sliding window takes 3 cases to train a classifier. The window is updated when 2 newly finished cases are inserted which satisfies retraining interval size. Window 1 contains case1 to case3. The window is updated after two new cases, case4 and case5, are inserted into a window. After updating the window, we retrain the model and use the model for detecting anomalies in newly arrived events (e.g., case6 and case7).

\section{Empirical evaluation}
\label{sec:empirical_evaluation}
The objectives of the evaluation are twofold. Firstly, we investigate how well our approach performs to detect anomalies in comparison to classical unsupervised anomaly detection as baseline. Secondly, we investigate the influence of the parameters on the detection performance, specifically the anomaly threshold, sliding window, and retraining interval. For these objectives, we implemented the proposed approach and the experiments in Python. The code to reproduce the experiments is publicly available on Github\footnote{ \url{https://github.com/ghksdl6025/streaming_anomaly_detect}}.
In the following, we first explain the experiment settings, which include the dataset used, the chosen techniques, and the parameter settings. Subsequently, we discuss the results of our evaluation. 

% %%%%%%%%%%%%%%%%%%%%%%%%%%%%%%%%%%%%%
% Experiment settings %
% %%%%%%%%%%%%%%%%%%%%%%%%%%%%%%%%%%%%%

\subsection{Setup of evaluation}
\label{sec:experiment_settings}
We have used synthetic logs with anomalous events that were used in~\cite{van2018filtering}. The event logs are generated from the gathered event streams based on the 21 variations of the loan application process~\cite{dumas2013fundamentals}. Then 6 anomalous event logs are created by randomly injecting the infrequently occurred events in various probability~\cite{van2018filtering}. Each log follows a different probability range from 2.5\% to 15\% in steps of 2.5\%. In this paper, the probability of generated anomalous events is denoted as \textit{Noise level}. Each log comprises 500 cases and approximately 7600 to 8600 events depends on the noise levels. Every log has 18 different activity labels. Table~\ref{tab:eventlog_statistics} shows descriptive statistics of the used event logs. 

\begin{table}[h]
\centering
\caption{Descriptive statistics of the noise imputed event log}
\label{tab:eventlog_statistics}
\begin{tabular}{c|c|c|c}
\hline
\textbf{Noise level} & \textbf{Cases} & \textbf{Events} & \textbf{Activity labels} \\ \hline
2.50\% & 500 & 7410 & 18 \\ \hline
5\% & 500 & 7630 & 18 \\ \hline
7.50\% & 500 & 8542 & 18 \\ \hline
10\% & 500 & 7922 & 18 \\ \hline
12.50\% & 500 & 7888 & 18 \\ \hline
15\% & 500 & 8159 & 18 \\ \hline
\end{tabular}
\end{table}

\begin{figure}[]
    \centering
    \includegraphics[width=0.75\linewidth]{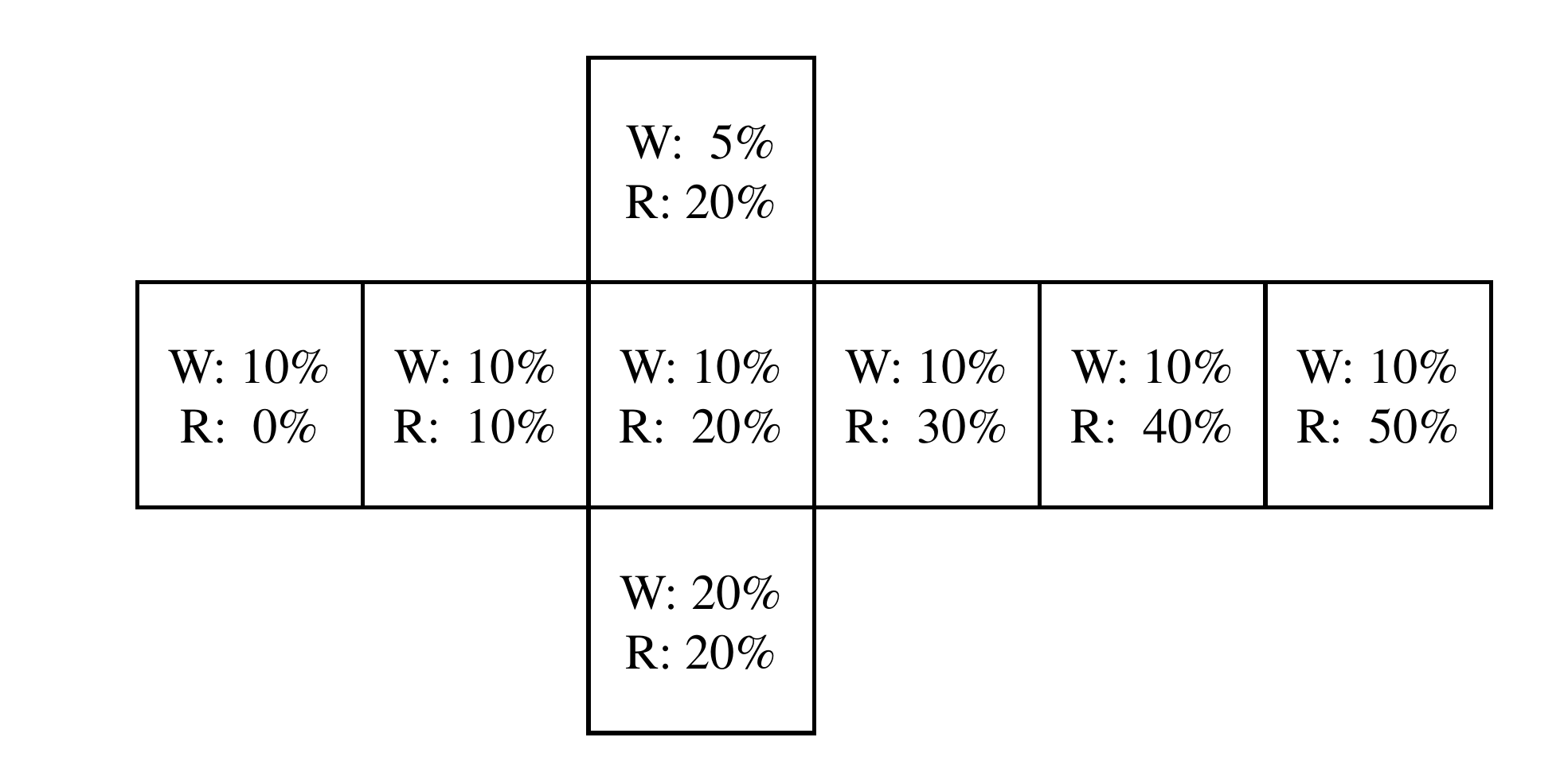}
    \caption{Evaluation settings. \textbf{W} indicates the percentage of total number of cases and \textbf{R} the percentage of sliding window}
    \label{fig:parameters}
\end{figure}

As discussed in Section~\ref{sec:methodology}, sliding window and retraining interval are parameters to respectively control the size of the training data and the retraining rate in the streaming setting. Fig.~\ref{fig:parameters} shows evaluation settings to check the influence of the sliding window and the retraining interval size. We consider 5\%, 10\%, and 20\% as a ratio to the total number of cases in the log as sliding window size. For the retraining interval, we tested with 6 different parameters from 0\% to 50\% in steps of 10\% as a ratio of retraining interval size to the sliding window. The influence of the sliding window is examined through changing sliding window sizes with fixed retraining interval, and vice versa. The threshold level is a proportion of anomalies in the dataset, which is used as a criteria in the anomaly detection phase. We experiment with 6 different thresholds, which have a range from 0.01 to 0.25, to investigate the influence of threshold on detection performance. 

In this evaluation, we consider both a machine learning model and a deep neural network to classify anomalous events, which are typically adopted in data mining field. As a detection model for the proposed approach, we experiment with two machine learning models and one deep neural network typically adopted in next activity prediction~\cite{teinemaa2019outcome}: random forest (RF), extreme gradient boosting (XGB), and long short term memory (LSTM). For  the baseline, we experiment with isolation forest (IForest), local outlier factor (LOF), one-class support vector machine (OCSVM), and autoencoder (AE), which are unsupervised anomalous data detection algorithms. For machine learning models, we use the classifiers provided by the Python packages Scikit-learn~\cite{pedregosa2011scikit}. 

For the deep neural network based detection model, the experiment is conducted with the Python package Pytorch~\cite{paszke2019pytorch} with different layer sizes and a number of layers by LSTM and AE. Regarding the LSTM model for the proposed method, two LSTM layers are stacked to obtain hidden feature vectors. Multiple linear layers are followed to predict next activity labels. We implement autoencoder structure including multi-linear layers for latent variables as presented in~\cite{nguyen2019autoencoders}. 

To performance of the anomaly detection task is separately evaluated using the F-score, ranged from 0 to 1, for both normal and anomalous events. The F-score is a classification model accuracy indicator calculated from precision and recall, i.e., correct rate among positive predictions and correct decision rate among true items, respectively. The high f-score, close to 1, indicates accurate identification of the events.

% %%%%%%%%%%%%%%%%%%%%%%%%%%%%%%%%%%%%%
% Detail step of unsupervised anomaly detection %
% %%%%%%%%%%%%%%%%%%%%%%%%%%%%%%%%%%%%%
\subsection{Baseline - Unsupervised Anomalous Event Detection}

\begin{figure}[]
        \centering
        \includegraphics[width=\linewidth]{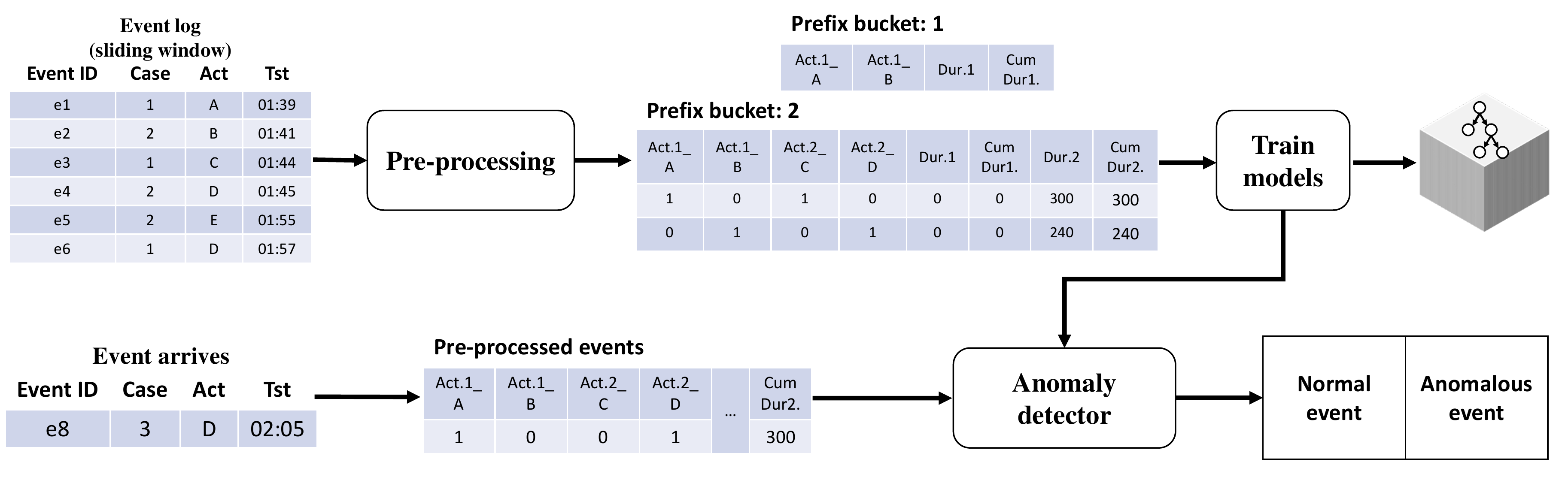}
        \caption{Online anomalous event detection procedure: Unsupervised method}
        \label{fig:framework_unsupervised}
\end{figure}

We propose to use classical unsupervised anomaly detection as the baseline approach. The approach is shown in Fig.~\ref{fig:framework_unsupervised}. We perform the same pre-processing step. However, the model detects the anomaly of an arrived event without predicting the next activity. This way of anomalous event detection is close to traditional outlier classification in data mining as explained in Section~\ref{sec:prelim.unsupervised}. The same pre-processing method is applied to encode input feature vector, as well as sliding window mechanism for streaming data. In the detection step, the classifier takes events of the running case with the target event as input and calculates deviated distance to the feature space. The model distinguishes the anomaly of an arrived event using an anomaly threshold.

\section{Result and discussion}
\label{sec:results}
% %%%%%%%%%%%%%%%%%%%%%%%%%%%%%%%%%%%%%
% Result and discussion %
% %%%%%%%%%%%%%%%%%%%%%%%%%%%%%%%%%%%%%

\begin{figure}[h]
    \centering
    \begin{subfigure}[]{\linewidth}
        \centering
        \includegraphics[width=0.85\linewidth]{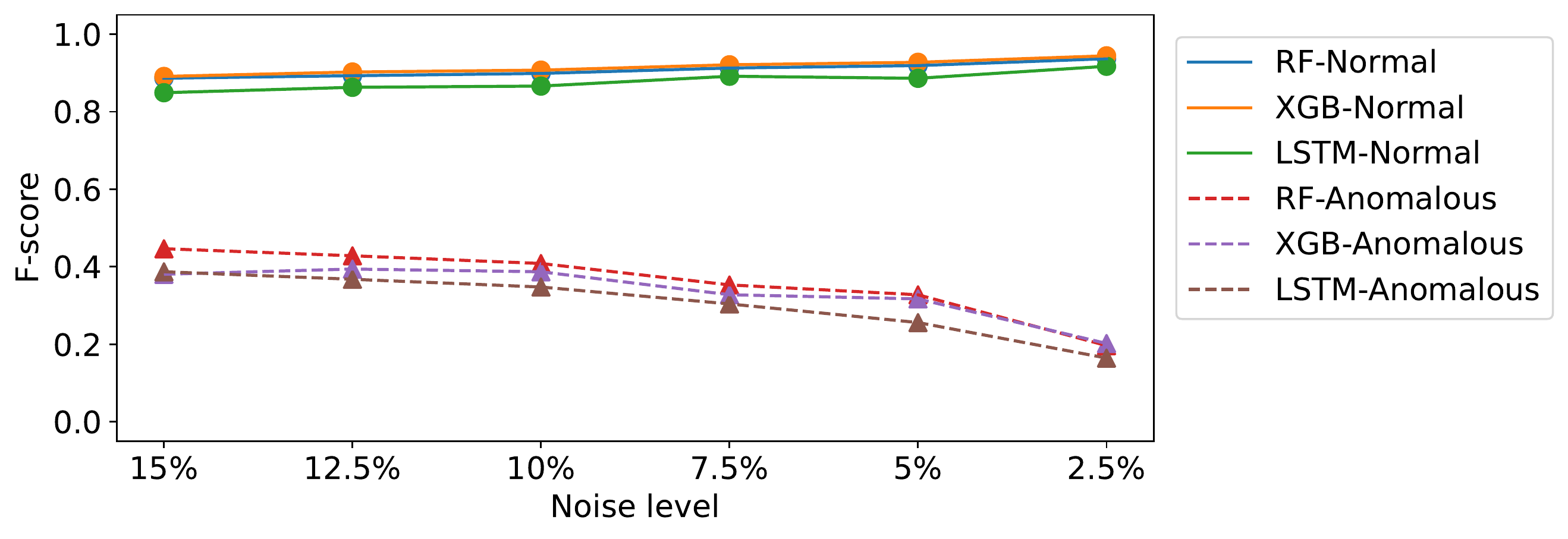}
        \caption{Proposed approach f-score}
        \label{fig:proposed_fscore}
    \end{subfigure}
        \hspace*{\fill}%  
    \begin{subfigure}[]{\linewidth}
        \centering
        \includegraphics[width=0.85\linewidth]{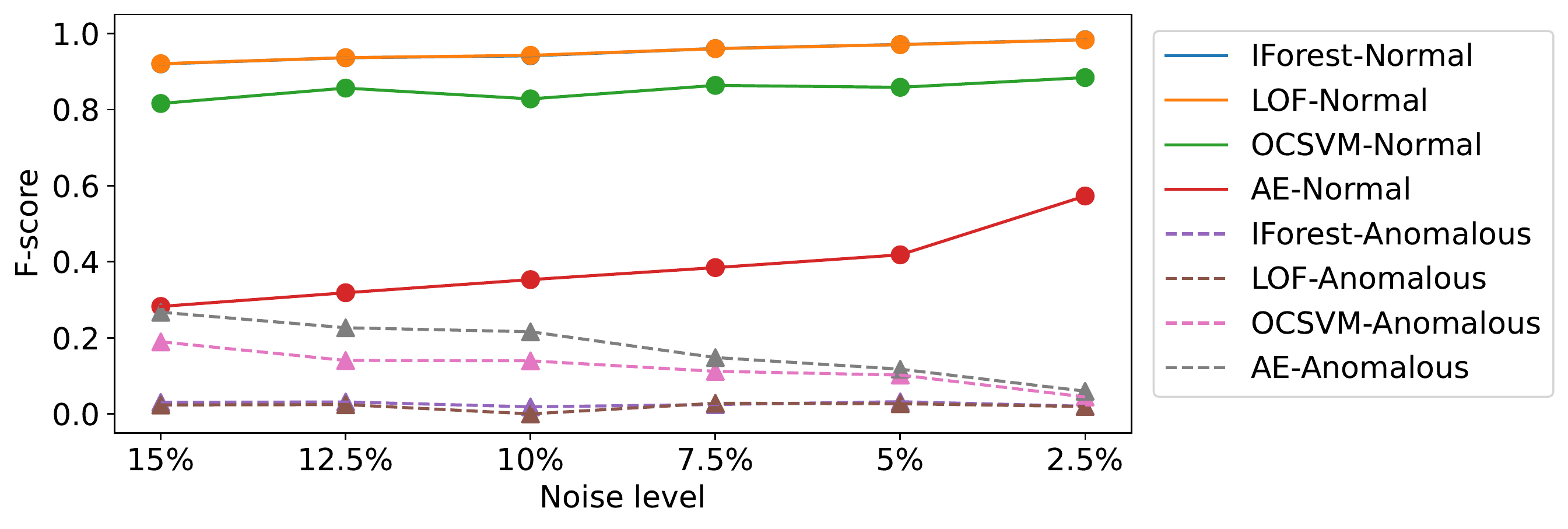}
        \caption{Unsupervised method f-score}
        \label{fig:unsupervised_fscore}
    \end{subfigure}
    \caption{F-score of normal and anomalous events detection by noise level}
    \label{Fig: prediction_framework}
\end{figure}

\subsubsection{Detection performance of proposed approach}

Fig.~\ref{Fig: prediction_framework} shows the F-score of normal and anomalous events detection for the proposed approach and the baseline. From the results, we find that (i) the proposed approach outperforms the baseline on every noise level and (ii) the deep neural network shows a lower performance than machine learning models. 

The proposed approach detects anomalous events more effectively than baseline for all every noise levels. Deep neural network based models in both approaches show lower performance than other classifiers. One possible reason for the performance gap is an issue on training data size. The sliding window size may not be big enough for proper deep learning model training.

% Figure.~\ref{fig:unsupervised_fscore} shows the f-score of outlier detection for unsupervised method. Except for relatively low performance in all noise levels of AE, f-score of anomalous event detection is lower than 0.2. In the point of comparison among machine learning based models, IForest and LOF show nearly 0 as f-score of anomaly detection, whereas two classifiers are better at selecting normal events. Two classifiers could not obtain anomalous event features which leads to allocating every event as normal one. On the other hand, OCSVM is possible to obtain information on anomalous event features.

\subsubsection{Effects of different anomaly threshold}

\begin{figure}[h]
    \centering
    \begin{subfigure}[]{\linewidth}
        \centering
        \includegraphics[width=0.9\linewidth]{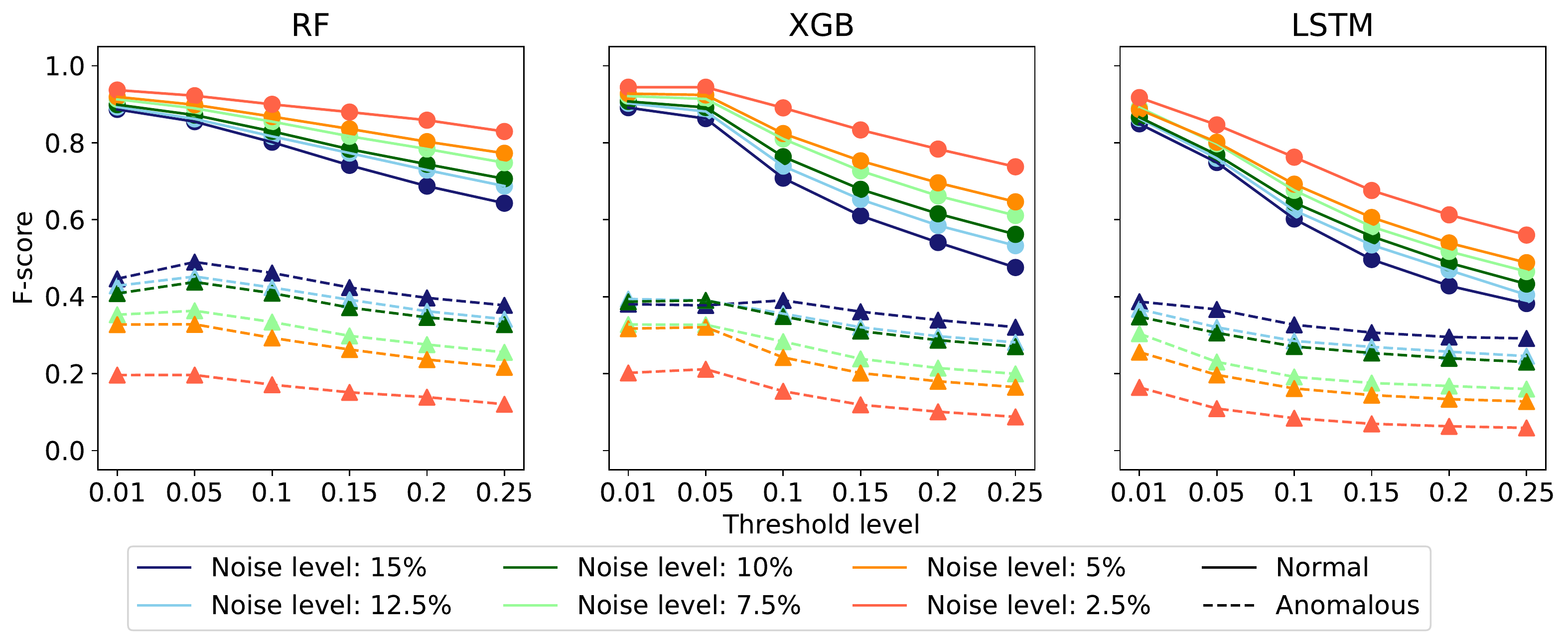}
        \caption{Proposed method}
        \label{fig:proposed_threshold}
    \end{subfigure}
        \hspace*{\fill}%  
    \begin{subfigure}[]{\linewidth}
        \centering

        \includegraphics[width=0.9\linewidth]{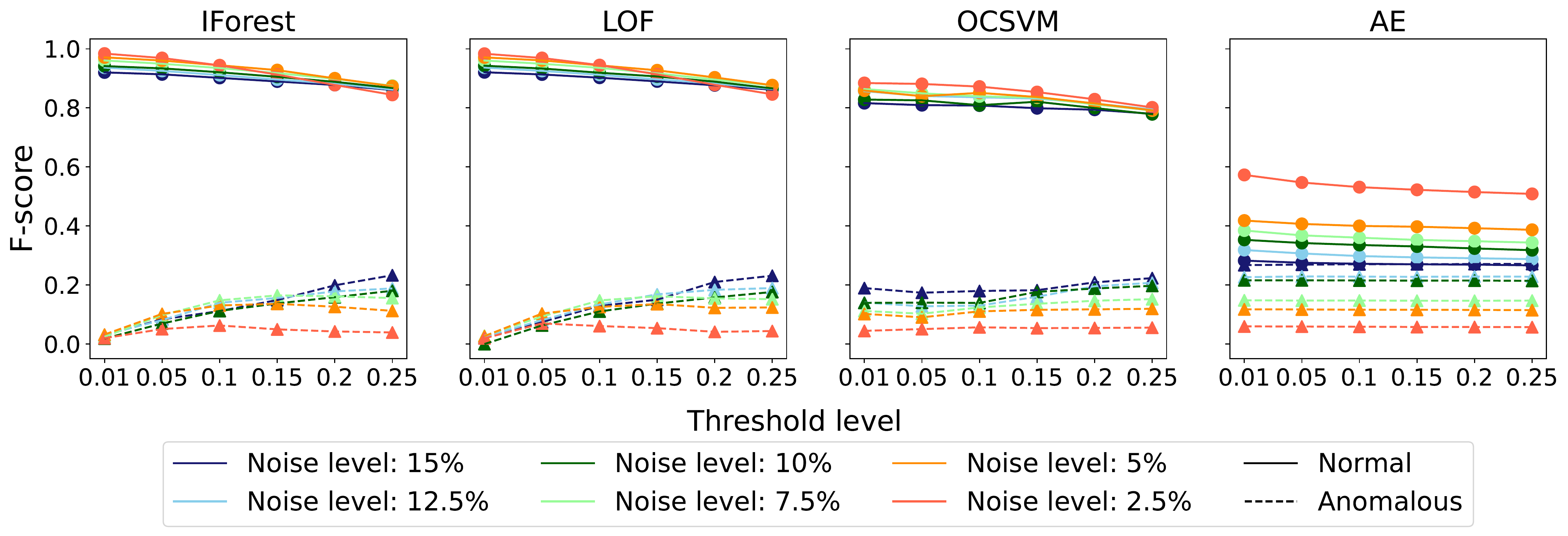}
        \caption{Unsupervised method}
        \label{fig:unsupervised_threshold}
    \end{subfigure}
    \caption{Performance comparison by anomaly threshold}
    \label{Fig: threshold_comparison}
\end{figure}

Fig.~\ref{Fig: threshold_comparison} shows the F-score of anomalous event detection for the proposed approach by different anomaly threshold levels. According to Fig.~\ref{fig:proposed_threshold}, the F-score of both normal and anomalous event decreases with the high threshold in next activity prediction, i.e., all classifiers perform the highest capability on detection at a 0.01 threshold level, except high noise level with RF. We can observe that each classifier requires a separate threshold level for optimal performance. Moreover, RF shows the highest performance at a 0.05 threshold with high noise event log, unlike the other classifiers. 

In case of the unsupervised approach, we observe two remarkable points from the analysis in Fig.~\ref{fig:unsupervised_threshold}. These are (i) different performance patterns among implemented algorithms and (ii) the influence of high threshold on anomalous event detection. Regarding performance volatility within models for unsupervised approach, the detection ability of IForest and LOF increases with high threshold while OCSVM and AE are relatively stable at all noise levels. In contrast to the first proposed approach, low threshold to distinguish anomalies could not be applied to the unsupervised method.

\subsubsection{Effects of sliding windows and retraining intervals}
\label{subsec:model_perf.param} We evaluate the anomaly detection performance of the proposed approach by different training windows and retraining interval sizes. 

\begin{figure}[h]
    \centering
    \begin{subfigure}[]{\linewidth}
        \centering
        \includegraphics[width=0.9\linewidth]{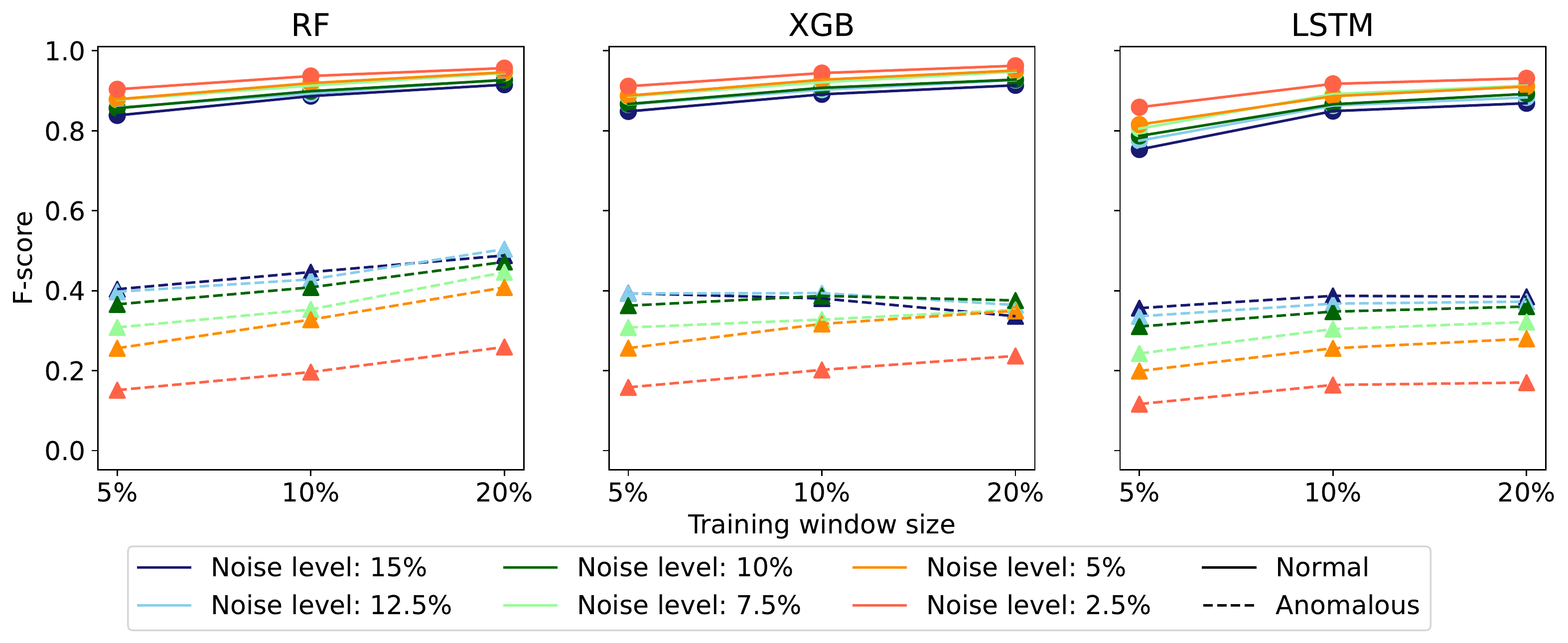}
        \caption{Proposed approach}
        \label{fig:proposed_window}
    \end{subfigure}
        \hspace*{\fill}%  
    \begin{subfigure}[]{\linewidth}
        \centering
        \includegraphics[width=0.9\linewidth]{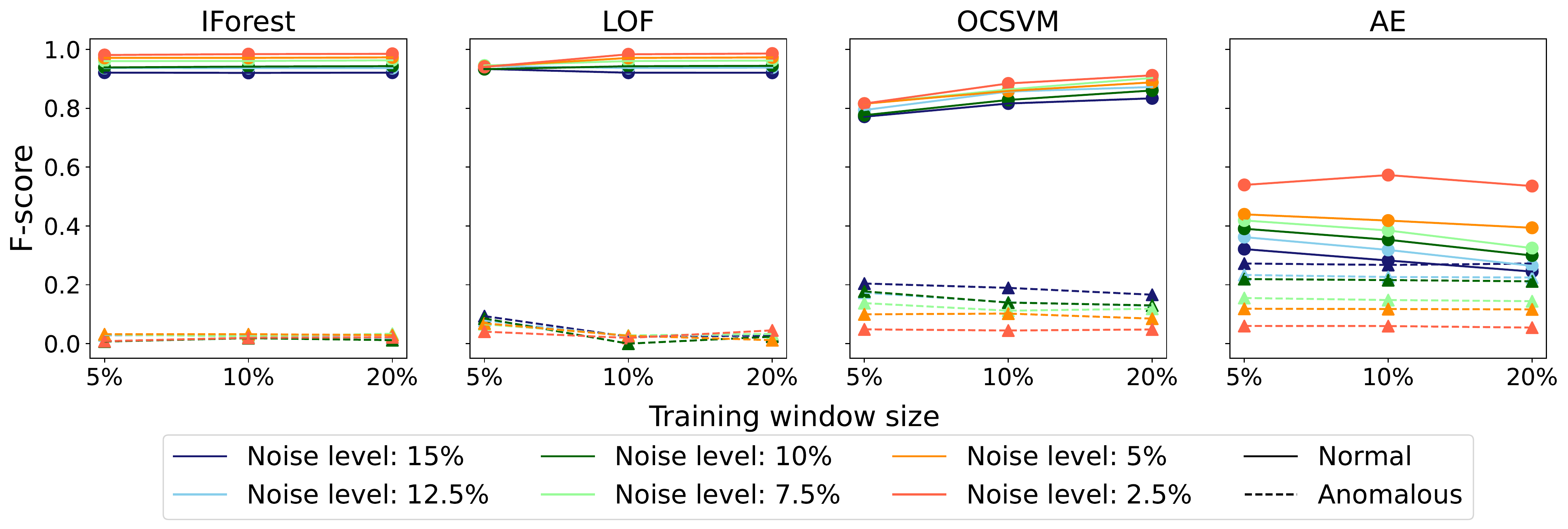}
        \caption{Unsupervised method}
        \label{fig:unsupervised_window}
    \end{subfigure}
    \caption{Performance comparison by sliding window size}
    \label{Fig: window_comparison}
\end{figure}

We find that (i) the sliding window size has a positive influence on better model performance, (ii) the proposed approach is more sensitive to window size change than the unsupervised approach, and (iii) the performance change slope is getting flattened with window size. Fig.~\ref{fig:proposed_window} and~\ref{fig:unsupervised_window} show the F-score of normal and anomalous event detection for the proposed approach and the unsupervised approach with sliding window change, respectively. Despite that, if both approaches were trained using the same window size, the sliding window size does not influence the performance improvement for the unsupervised approach. More specifically, the results regarding OCSVM seem to show that the increase of training window size even has a negative impact on detecting the anomalies.

The performance of the proposed approach improves for both normal and anomalous events detection with a large window, i.e., more useful information is collected with a bigger window size. Along with the positive relation between model improvements and training data size, we also observe a marginal effect of increasing training window and detecting ability improvements. As in Fig.~\ref{fig:proposed_window}, the slope between sliding window size and F-score is not linear, i.e., the obtained useful information for detecting anomalous by increasing training dataset is limited. Therefore, the efficiency of managing the training window depends on a balance between information gain and the cost of increasing the dataset. 

\begin{figure}[h]
    \centering
    \begin{subfigure}[]{\linewidth}
        \centering
        \includegraphics[width=0.9\linewidth]{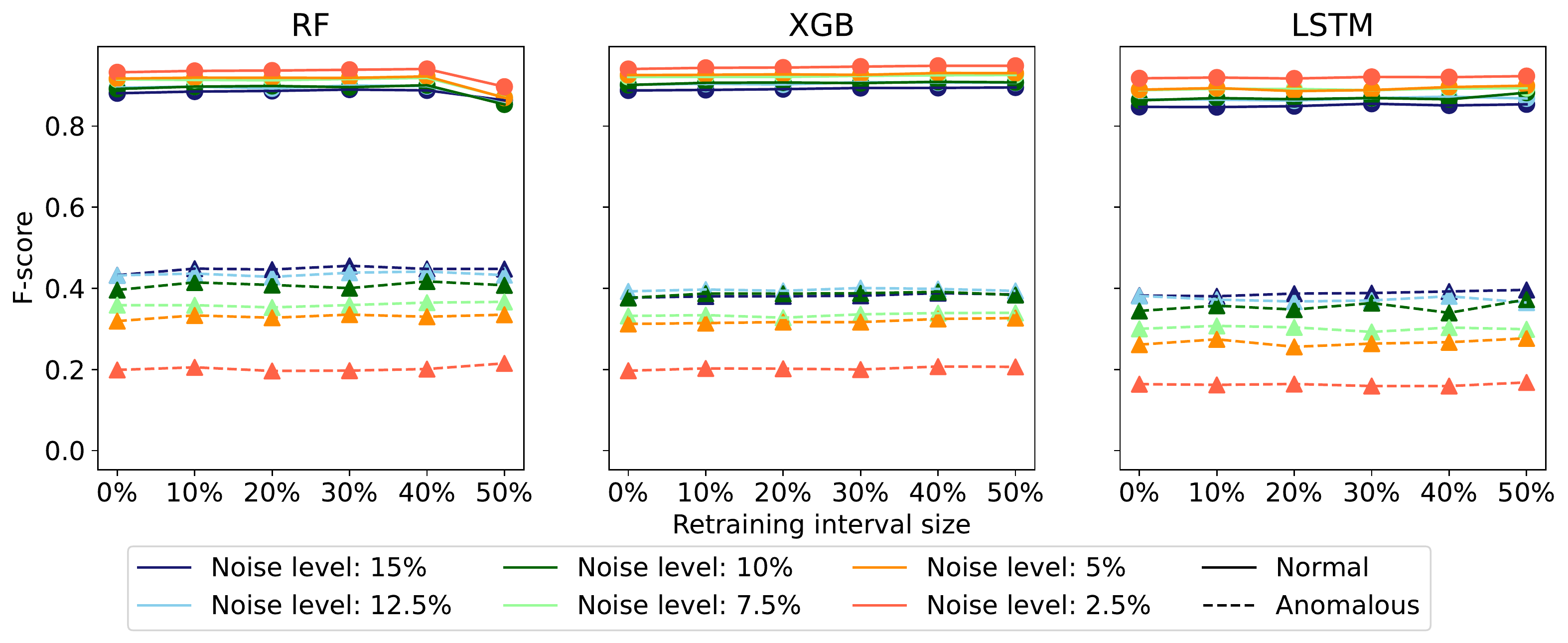}
        \caption{Proposed method}
        \label{fig:proposed_retraining}
    \end{subfigure}
        \hspace*{\fill}%  
    \begin{subfigure}[]{\linewidth}
        \centering
        \includegraphics[width=0.9\linewidth]{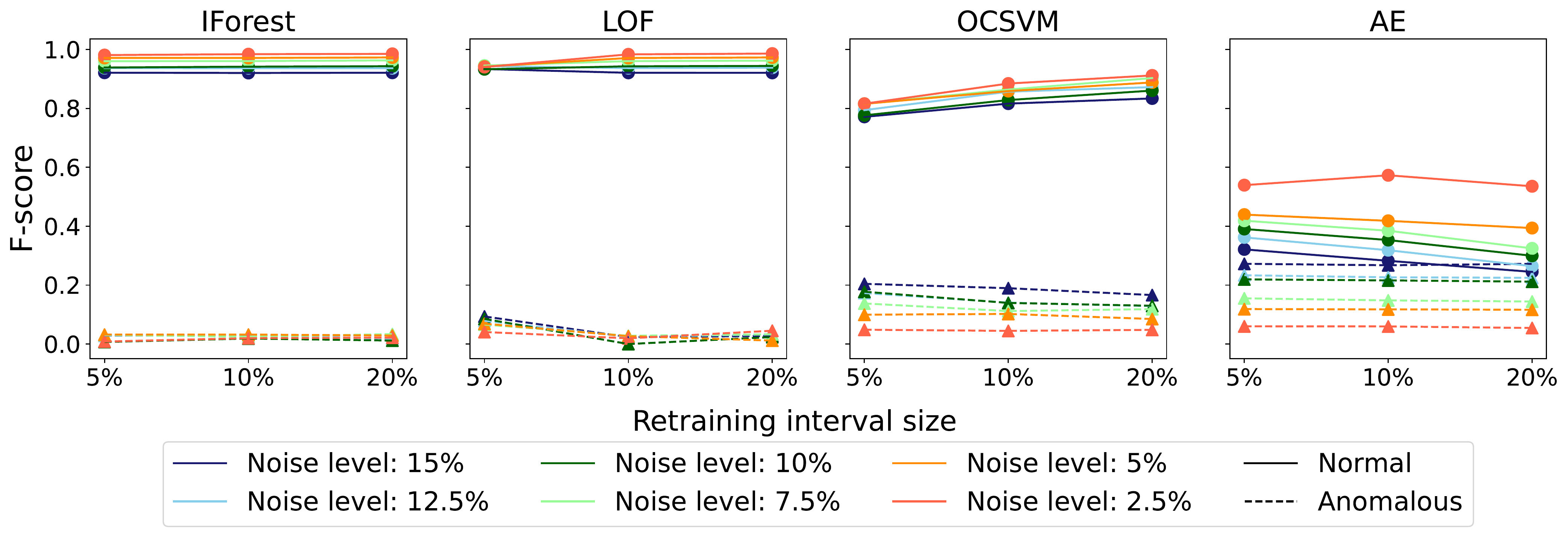}
        \caption{Unsupervised method}
        \label{fig:unsupervised_retraining}
    \end{subfigure}
    \caption{Performance comparison by retraining interval size}
    \label{Fig: retraining_comparison}
\end{figure}

Retraining interval size does not influence the anomaly detection ability for either the proposed approach or unsupervised method. Fig.~\ref{fig:proposed_retraining} and~\ref{fig:unsupervised_retraining} show the F-score of normal and anomalous event detection for the proposed approach and the unsupervised algorithm with retraining interval changes, respectively. Generally, we can observe that the retraining interval is irrelevant to the model training phase. The performance lines across all detecting models are stable over noise level.

% \section{Research challenge}
% \label{sec:research_challenge}
% \input{files/research_challenge}

\section{Conclusion and challenges}
\label{sec:conclusion}
We have presented an approach for online event anomaly detection via next activity prediction using machine learning models. The proposed approach uses a sliding window to (re)train the model with the recently recorded cases on the event stream. The probability of possible next activities obtained from the ML models is used to classify a new event as anomalous or not. 

The method uses well-known machine learning and deep learning algorithms, which can give flexibility by changing into other models later. Even though the performance of deep neural networks is lower than machine learning models, we have shown that the next activity prediction method outperforms the classical unsupervised anomaly detection method when applied to event logs.

As a relatively new research area, anomalous event detection in event streams has open challenges in several perspectives. We identify multiple challenges on online anomaly detection that still need to be addressed.
Firstly, in an online setting, an event classified as (potential) anomalous may be changed into normal behavior later after the model update or after new events arrives.\emph{ (\textbf{RC1}) How and when can we confirm that a potential anomalous event is a definitive anomaly?} The approach proposed in this paper is designed to not updating the prediction. Nevertheless, taking such changes into account may be very informative for suggesting user actions and is a challenge itself and worthy of further investigation. Another interesting challenge related to changes is
\emph{ (\textbf{RC2}) how can anomalies be detected while taking into account concept drift? }
Finally, the online detection of potential anomalies leads to the possibility of building an online decision-making system that suggests follow-up actions for the detected anomalies. This possibility leads to the third research challenge: \emph{(\textbf{RC3}) how can we build a decision-making system with the domain experts for the further investigation of the potential anomalies? } 
    % \xixi{The same as above}
% \end{enumerate}

% \begin{enumerate}[label=]%[label=\textbf{RQ{\arabic*}}]
%     \item \RQ{RC1}{How can we confirm the potential anomalous event as definite anomalies with respect to the updating model and decision?}
%     \item \RQ{RC2}{How can we guarantee case termination in a streaming event log?} 
%     % \xixi{What do you mean with case initialization and termination?}
%     \item \RQ{RC3}{How can anomalies in an event log can be detected taking into account concept drift occurrence?}
%     % \xixi{It is a nice challenge but did your approach solve this? And do you have evidence that you approach solves this challenge?}
%     % \item \RQ{RC4}{How can we build a decision-making system with the domain experts for the second investigation on the potential anomalies? }
%     % \xixi{The same as above}
% \end{enumerate}

The work presented here can be extended in several ways. In addition to solving the research challenges, we will study the post-hoc analysis on the online anomaly detecting model to provide an explanation of the cause of anomalous events. Finally, we are planning to develop the interactive online detecting model by implementing the feedback from users who select anomalous events. The collected feedback would help the model to give concrete reasoning on the predicted anomalies and improve the model performance.

% ---- Bibliography ----
%
% BibTeX users should specify bibliography style 'splncs04'.
% References will then be sorted and formatted in the correct style.
%
% \bibliographystyle{splncs04}
% \bibliography{mybibliography}
%
\bibliographystyle{splncs04}
\bibliography{mybiblio}

\end{document}